\documentclass[conference]{IEEEtran}

\usepackage{cite}
\usepackage{flushend}
\usepackage{amsmath,amssymb,amsfonts}
\usepackage{algorithmic}
\usepackage{multirow}
\usepackage{booktabs}
\usepackage{subcaption}
\usepackage{graphicx}
\usepackage{hyperref}
\graphicspath{ {./images/} }
\usepackage[braket, qm]{qcircuit}
\usepackage{textcomp}
\usepackage[table]{xcolor}
\usepackage{dirtytalk}
\usepackage{tikz}
\usepackage{url} 

\usepackage{caption}
\usetikzlibrary{positioning}
\def\BibTeX{{\rm B\kern-.05em{\sc i\kern-.025em b}\kern-.08em
    T\kern-.1667em\lower.7ex\hbox{E}\kern-.125emX}}
    
\begin{document}



\title{Numerical Instability and Chaos: Quantifying the Unpredictability of Large Language Models}

\author{
\IEEEauthorblockN{Chashi Mahiul Islam$^1$, Alan Villarreal$^1$, Mao Nishino$^2$, Shaeke Salman$^1$, Xiuwen Liu$^1$}
\IEEEauthorblockA{$^1$\textit{Department of Computer Science, Florida State University, Tallahassee, USA} \\
$^2$\textit{Department of Mathematics, Florida State University, Tallahassee, USA} \\
ci20l@fsu.edu, adv23a@fsu.edu, mnishino@fsu.edu, salman@cs.fsu.edu, xliu@fsu.edu}
}

\maketitle

\begin{abstract}
As Large Language Models (LLMs) are increasingly integrated into agentic workflows, their unpredictability stemming from numerical instability has emerged as a critical reliability issue. While recent studies have demonstrated the significant downstream effects of these instabilities, the root causes and underlying mechanisms remain poorly understood. In this paper, we present a rigorous analysis of how unpredictability is rooted in the finite numerical precision of floating-point representations, tracking how rounding errors propagate, amplify, or dissipate through Transformer computation layers. Specifically, we identify a chaotic ``avalanche effect"  in the early layers, where minor perturbations trigger binary outcomes: either rapid amplification or complete attenuation. Beyond specific error instances, we demonstrate that LLMs exhibit universal, scale-dependent chaotic behaviors characterized by three distinct regimes: 1) a stable regime, where perturbations fall below an input-dependent threshold and vanish, resulting in constant outputs; 2) a chaotic regime, where rounding errors dominate and drive output divergence; and 3) a signal-dominated regime, where true input variations override numerical noise. We validate these findings extensively across multiple datasets and model architectures.
\end{abstract}



\section{Introduction}

The deployment of Large Language Models (LLMs) has rapidly evolved from isolated inference systems to complex, distributed multi-agent architectures where multiple AI agents collaborate to solve sophisticated tasks~\cite{wang2024survey, xi2023rise}. Multi-agent LLM systems improve performance through collaborative problem-solving~\cite{chen2024agentverse, wu2023autogenenablingnextgenllm}, but they exhibit high failure rates that cannot be attributed solely to algorithmic limitations. Wu et al.~\cite{wu2023autogenenablingnextgenllm} report that AutoGen-based workflows fail to converge on 23\% of collaborative tasks, with agents producing contradictory outputs despite identical prompts. Similarly, Hong et al.~\cite{hong2024metagpt} document non-reproducible outputs in 31\% of MetaGPT planning tasks across identical hardware configurations. These failures share a common signature: unpredictability that persists even with fixed random seeds, manifesting most severely when agents exchange intermediate representations.

We hypothesize that a significant fraction of these multi-agent failures stem from numerical instability induced by floating-point computation across heterogeneous infrastructure. In modern LLM deployments, floating-point arithmetic is neither associative nor deterministic across diverse hardware~\cite{goldberg1991floating}. When agents communicate by exchanging embeddings across cloud environments with diverse GPU architectures, numerical discrepancies arise from non-deterministic parallel reduction orders~\cite{demmel2013parallel}, hardware-specific implementations~\cite{nvidia2024cudnn}, and accumulation of rounding errors through deep network layers. The Oak Ridge National Laboratory documents that ``the variability from non-deterministic reductions will produce unique and non-reproducible results with each run using identical inputs''~\cite{shanmugavelu2024impacts}. Zhuang et al.~\cite{zhuang2022randomness} demonstrate that GPU-based training produces non-reproducible results in 100\% of trials across different hardware.

This impact is amplified by the extreme sensitivity of LLMs. Hahn~\cite{hahn2020theoretical} proves that self-attention mechanisms have condition numbers that grow exponentially with sequence length. Kim et al.~\cite{kim2024persona} demonstrate empirically that LLMs are ``extremely sensitive to assigned prompts,'' with 13.78\% of correct answers becoming incorrect solely due to minor prompt perturbations. This sensitivity extends to embeddings: Vijayaraj~\cite{vijayaraj2025embeddingsdiagnosislatentfragility} shows that perturbations of magnitude $10^{-3}$ in normalized embedding space cause diagnostic prediction flips in 12.4\% of clinical cases.

Prior work has documented numerical instability in deep learning, and multi-agent system fragility~\cite{shanmugavelu2024impacts, zhuang2022randomness,wu2023autogenenablingnextgenllm, guo2024large}, but a gap remains: \textit{we lack a principled understanding of how floating-point rounding errors interact with LLM computation to produce unpredictable outputs}. Existing work treats numerical instability as an engineering nuisance to be mitigated through deterministic execution modes~\cite{pytorch2024reproducibility} or higher precision arithmetic, without characterizing the underlying dynamics.

This paper addresses this gap through rigorous analysis of numerical stability in LLM inference. Our contributions are:

\begin{enumerate}
    \item \textbf{Identification of Chaotic Dynamics.} We demonstrate that LLMs exhibit chaotic behavior where perturbations at the scale of floating-point epsilon ($\sim 10^{-14}$) either amplify exponentially or attenuate completely within early Transformer layers, with directional absolute condition numbers exceeding $10^6$.
    
    \item \textbf{Characterization of Three Stability Regimes.} We identify three distinct operating regimes: \textit{Constant Regions} where output is bitwise-constant; \textit{Chaotic Regions} with extreme sensitivity; and \textit{Signal-Dominated Regions} where input variations override numerical noise.
    
    \item \textbf{Empirical Validation.} We validate findings across multiple combinations of settings, including architectures (Llama-3.1-8B, GPT-OSS-20B), datasets (TruthfulQA, AdvBench), and floating-point precision (BFloat16, FP32, and FP64), demonstrating universal phenomena rather than model-specific artifacts.
\end{enumerate}

As LLMs are integrated into safety-critical applications~\cite{singhal2023large, wu2023autogenenablingnextgenllm}, understanding boundaries between reliable and chaotic operation becomes essential. Our characterization provides practitioners with principled guidance for building robust multi-agent systems.



\section{Related Work}

\subsection{Numerical Stability in Deep Learning}
Goldberg~\cite{goldberg1991floating} established that floating-point operations are neither associative nor commutative, while Demmel and Nguyen~\cite{demmel2013parallel} proved parallel reductions are inherently non-deterministic. Recent work documented these issues in deep learning: Zhuang et al.~\cite{zhuang2022randomness} showed non-reproducible results across identical seeds, Nagarajan et al.~\cite{nagarajan2019deterministicimplementationsreproducibilitydeep} quantified 40\%+ variance in deep RL, and Shanmugavelu et al.~\cite{shanmugavelu2024impacts} demonstrated non-deterministic weight evolution. Yuan et al.~\cite{yuan2025understanding} systematically characterized numerical sources of nondeterminism in LLM inference and proposed mitigation strategies. While quantization studies~\cite{dettmers2023qlora,frantar2022gptq,xiao2023smoothquant} extensively analyzed reduced-precision arithmetic, we examine \textit{inherent instability at full Float32 precision}.

\subsection{Sensitivity and Chaos in Neural Networks}
Hahn~\cite{hahn2020theoretical} proved Transformer condition numbers grow exponentially with sequence length. Recent work explored chaos in neural networks: Liu et al.~\cite{liu2024exploiting} deliberately exploited chaotic dynamics for improved performance, Terada and Toyoizumi~\cite{terada2024chaotic} showed chaos facilitates Bayesian sampling in RNNs, and Engelken et al.~\cite{engelken2023lyapunov} computed Lyapunov spectra for recurrent networks. However, these focus on \textit{beneficial} chaos or RNN-specific dynamics, not unintended chaos from floating-point errors in Transformers. For adversarial robustness, Wallace et al.~\cite{wallace2019universal} discovered universal triggers and Zou et al.~\cite{zou2023universal} achieved 80\%+ jailbreak rates, but these involve semantic-level perturbations rather than numerical instabilities.

\subsection{LLM Robustness and Multi-Agent Systems}
Kim et al.~\cite{kim2024persona} showed 13.78\% of correct answers flip due to persona changes, while Vijayaraj~\cite{vijayaraj2025embeddingsdiagnosislatentfragility} found $10^{-3}$ magnitude perturbations cause 12.4\% diagnostic flips. Tuck et al.~\cite{tuck2025assessing} measured embedding sensitivity for adversarial detection, but focused on semantic perturbations. In multi-agent systems, Wu et al.~\cite{wu2023autogenenablingnextgenllm} reported 23\% AutoGen failure rates and Hong et al.~\cite{hong2024metagpt} documented 31\% non-reproducible MetaGPT outputs. Guo et al.~\cite{guo2024large} identified brittle communication behaviors but attributed failures to algorithmic factors rather than numerical instability.

Unlike quantization studies that examine reduced precision~\cite{dettmers2023qlora,frantar2022gptq} or work deliberately exploiting chaos~\cite{terada2024chaotic,liu2024exploiting}, we characterize \textbf{unintended chaotic behaviors from floating-point rounding at Float32}. While prior sensitivity analyses~\cite{hahn2020theoretical,tuck2025assessing} focus on theoretical bounds or semantic perturbations, and Yuan et al.~\cite{yuan2025understanding} address determinism in inference pipelines, we provide the \textbf{first systematic layer-wise analysis showing machine-epsilon perturbations ($\sim 10^{-14}$) either amplify to around $10^{-6}\times$ or vanish, defining three operating regimes} (Constant, Chaotic, Signal-Dominated) with empirical validation across multiple LLM architectures.

\section{Methodology}

To quantify the stability of an LLM with respect to specific input perturbations, we adopt a directional approach~\cite{Nesterov_Coordinate_2012}. While the standard condition number typically denotes the worst-case sensitivity (the spectral norm of the Jacobian)~\cite{higham2002accuracy}, this metric is often overly pessimistic for high-dimensional neural networks, such as LLMs. 

Instead, we utilize the norm of the directional derivative,\footnote{Mathematically, there are two types of directional derivative: G\^{a}teaux derivative and Fr\'echet derivative. The differences do not change the final solution here.} which we refer to as the \textbf{absolute directional condition number}. Following the foundational theory of conditioning by Rice \cite{rice1966theory}, the absolute condition number is the limit of the error magnification factor. By restricting this definition to a specific perturbation direction $v$, we obtain a precise measure of local stability.

Let $f: \mathbb{R}^n \to \mathbb{R}^m$ represent the LLM mapping from the embedding space to the output space. For an input $x \in \mathbb{R}^n$ and a normalized perturbation direction $v \in \mathbb{R}^n$ (where $\|v\|_2 = 1$), the absolute directional condition number $\kappa_{abs}(f, x, v)$ is defined as:

\begin{equation}
    \kappa_{abs}(f, x, v) := \| L_f(x; v) \|_2 = \| J(x) v \|_2
\end{equation}

\noindent where $L_f(x; v)$ denotes the directional derivative of $f$ at $x$ in the direction $v$ \cite{higham2008functions}, and $J(x)$ is the Jacobian matrix. Numerically, this estimates the immediate absolute change in the output given a unit perturbation along $v$:

\begin{equation}
    \kappa_{abs}(f, x, v) \approx \frac{\| f(x + \epsilon v) - f(x) \|_2}{\epsilon} \quad \text{for } \epsilon \to 0.
\end{equation}
A value of $\kappa_{abs} \gg 1$ indicates that the model is numerically unstable in the direction $v$, expanding small input noises into significant output deviations.

A challenge applying $\kappa_{abs}$ to LLMs is that their final outputs are probabilistic in nature, which depend on the choice of hyperparameters (such as temperature) and the sampling algorithm to be used. To overcome the issue, we use the output before the last language modeling head layer. 
To define it unambiguously, we decompose the LLM computation into two parts, as given
\begin{equation}
    P(y \mid x) = \text{softmax}(\underbrace{W_U \cdot f(x)}_{\text{Logits } z})
\end{equation}

\noindent where:
\begin{itemize}
    \item $f(x) \in \mathbb{R}^{d_{model}}$ denotes the final hidden state from the Transformer backbone, which is named as the last pseudo token (LST) in this paper as it is a continuous vector, not limited to the ones in the model's dictionary.
    \item $W_U \in \mathbb{R}^{V \times d_{model}}$ is the unembedding matrix (or output projection).
    \item $z = W_U \phi(x)$ represents the unnormalized logits.
\end{itemize}

As we show numerically, for LLMs, when estimated numerically, $\kappa_{abs}$ can be several orders of magnitude larger than the singular value of the Jacobian, especially for small $\epsilon$ along the directions where the singular values are small, indicating that the stability of the LLMs is affected by numerical issues that have not been explored. Furthermore, we show that the numerical estimation of $\kappa_{abs}$ converges to a value independent of the singular value of the direction before it vanishes. While the phenomenon is very important and has directly related to the issues reported in earlier studies such as~\cite{yuan2025understanding}, it has not been studied. We show the phenomenon is universal to LLMs.

\textbf{Experimental Setting: }

To rigorously validate these regions, we conducted experiments across different model architectures and datasets.

\textbf{Models and Hardware:} 

We evaluated \texttt{Meta-Llama-3.1-8B-Instruct} and \texttt{OpenAI-GPT-OSS-20B}. To capture precise floating-point behaviors, specific hardware configurations were required. \texttt{Llama-3.1-8B} experiments were conducted on dual NVIDIA RTX A5000 GPUs (24GB VRAM each). Due to memory constraints when enforcing Float32 precision on the larger model, \texttt{GPT-OSS-20B} experiments were executed on an Intel Core i9-10900X CPU.

\textbf{Datasets:}

We utilized \textit{TruthfulQA} to evaluate stability on general knowledge and reasoning tasks. Additionally, we employed \textit{AdvBench} (subset of harmful behaviors) to test stability on adversarial prompts, determining if safety guardrails introduce additional numerical instability.


\section{Results and Analysis}

\subsection{Directional sensitivity is largely epsilon-driven}

We define the \textbf{effective directional condition number} $D(\epsilon, v) := \|f(x + \epsilon v) - f(x)\|_2 / \epsilon$ as the finite-$\epsilon$ approximation to $\kappa_{abs}(f, x, v)$.

To determine whether directional sensitivity is determined by the Jacobian spectrum (as in classical local conditioning) or whether it is governed mainly by the perturbation scale $\epsilon$ in floating-point implementations, we analyze the effective directional number across multiple singular directions spanning the spectrum.

\begin{figure}[t]
    \centering
    \includegraphics[
        width=0.92\columnwidth,
        trim=0 0 0 25,
        clip
    ]{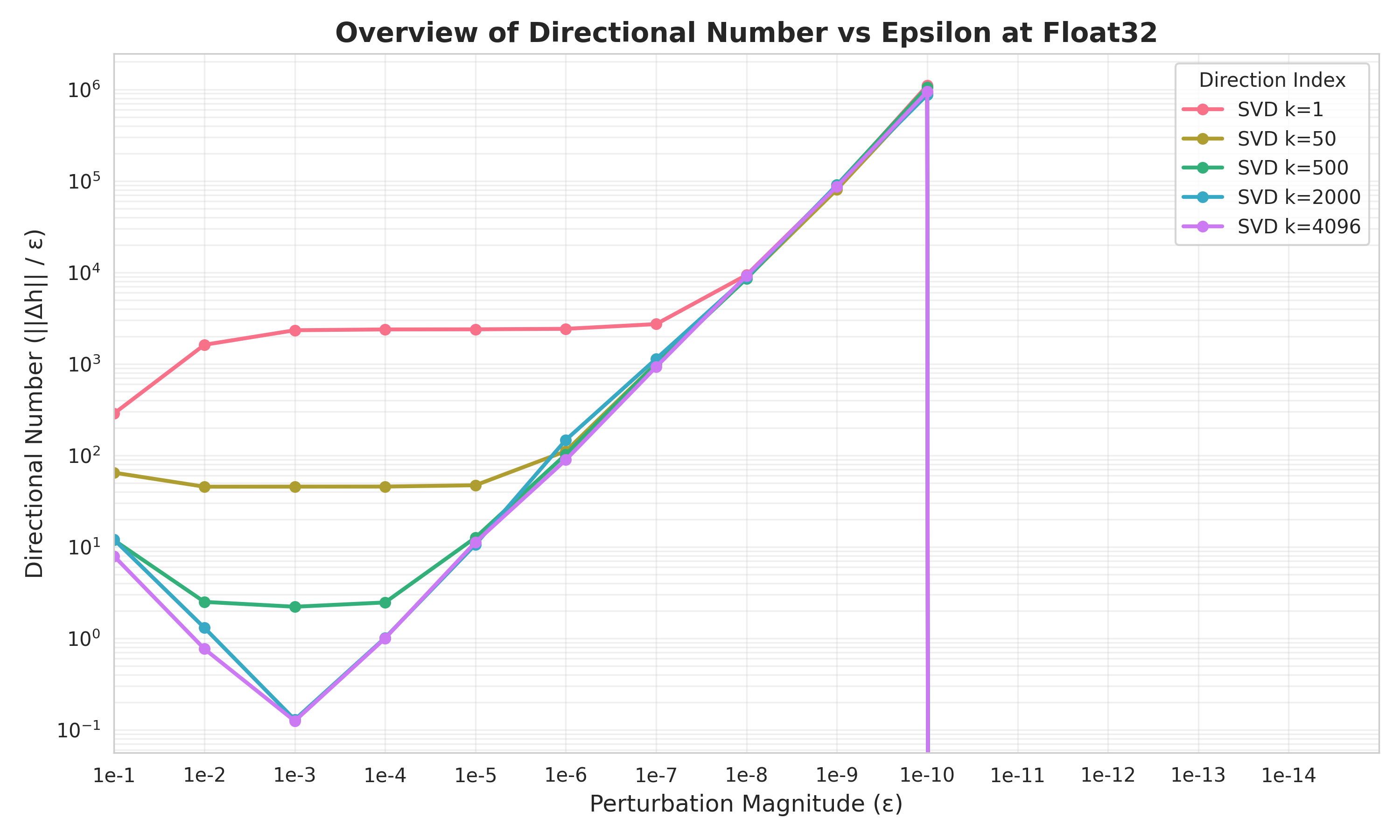}
    \caption{Overview of effective directional number $D(\epsilon,v)$ at layer index 32 (Float32) for $v_1, v_{50}, v_{500}, v_{2000}, v_{4096}$.}
    \label{fig:directional_overview_fp32}
\end{figure}

Figure~\ref{fig:directional_overview_fp32} plots the effective directional number $D(\epsilon,v)$ at the last representative layer (\texttt{Llama-3.1-8B}) for five singular directions spanning the spectrum ($v_1, v_{50}, v_{500}, v_{2000}, v_{4096}$). The curves exhibit the same qualitative scale dependence: at sufficiently small $\epsilon$, the response becomes dominated by representational granularity and finite-precision effects rather than by the singular value ranking. This is the sense in which the implementation's \emph{effective} directional sensitivity can be largely $\epsilon$-driven, even though the underlying Jacobian spectrum is well-defined.

This observation is not a claim about the ideal mathematical condition number (defined by a limit as $\epsilon \to 0$ for a real-valued function). Instead, it characterizes the behavior of the finite-precision program under finite perturbations, where the spacing between representable numbers (ULP) depends on magnitude and can create threshold effects as $\epsilon$ changes. \cite{goldberg1991floating} \cite{8766229}

\subsection{Layer-wise propagation shows spectrum collapse at small epsilon}
Sensitivity can either grow or vanish through network depth. If the model amplifies microscopic differences, layer-wise growth can overwhelm the initial directional structure. To separate these behaviors, we compare propagation at a large perturbation scale and a microscopic perturbation scale and identify when directional structure is preserved versus when it collapses.

\begin{figure}[t]
    \centering
    \includegraphics[width=0.92\columnwidth,
    trim=0 0 0 25,
        clip
    ]{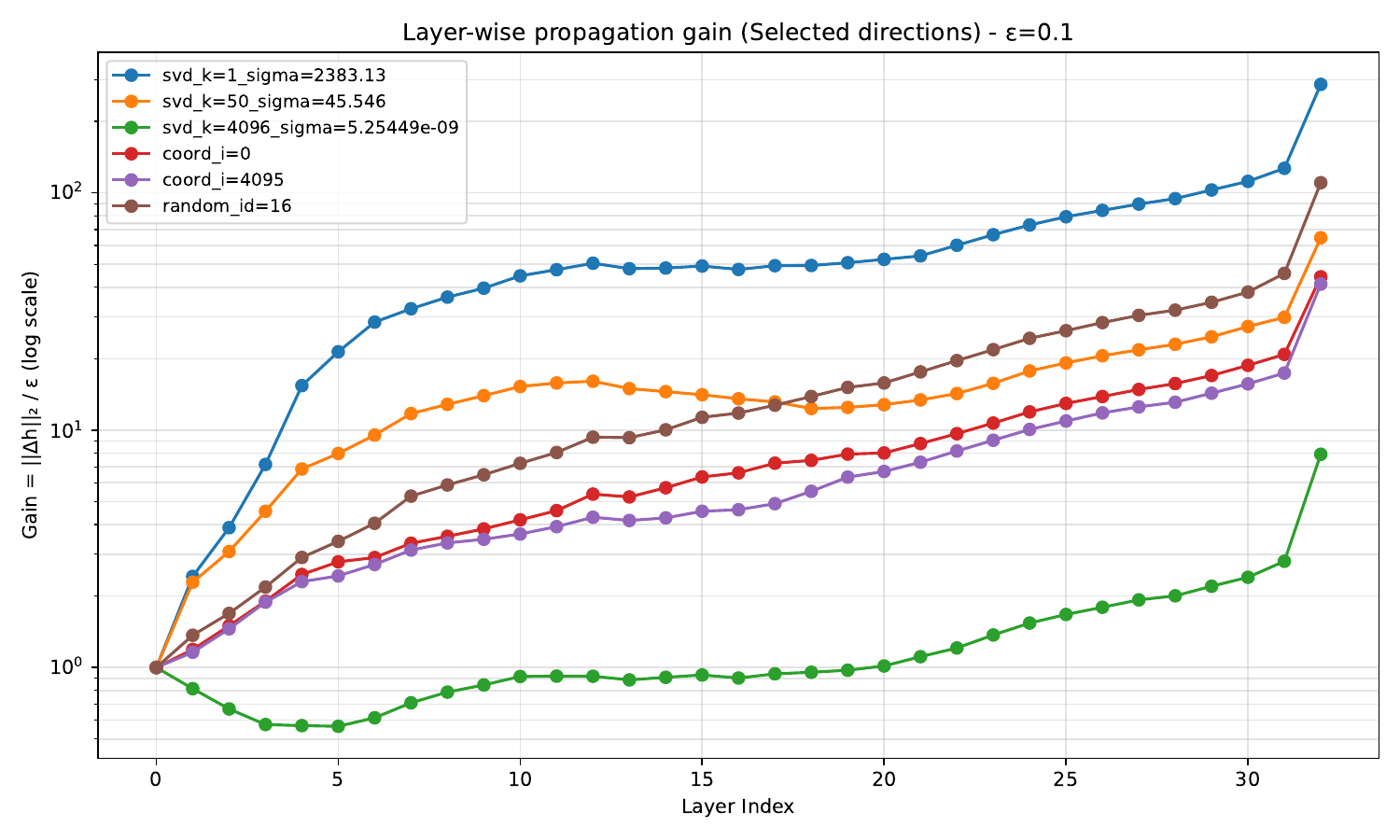}
    \caption{Layer-wise propagation profile at $\epsilon=0.1$. Directional structure is visible: higher-$\sigma$ directions yield larger amplification than low-$\sigma$ directions, consistent with a signal-dominated regime.}
    \label{fig:layer_gain_signal}
\end{figure}

Figure~\ref{fig:layer_gain_signal} shows the layer-wise propagation behavior at a large perturbation ($\epsilon=0.1$). In this regime, the ordering across directions broadly follows the singular spectrum: the top singular direction exhibits larger amplification while low-$\sigma$ directions remain comparatively suppressed, consistent with a signal-dominated behavior.

\begin{figure}[t]
    \centering
    \includegraphics[width=0.92\columnwidth,
    trim=0 0 0 25,
        clip
    ]{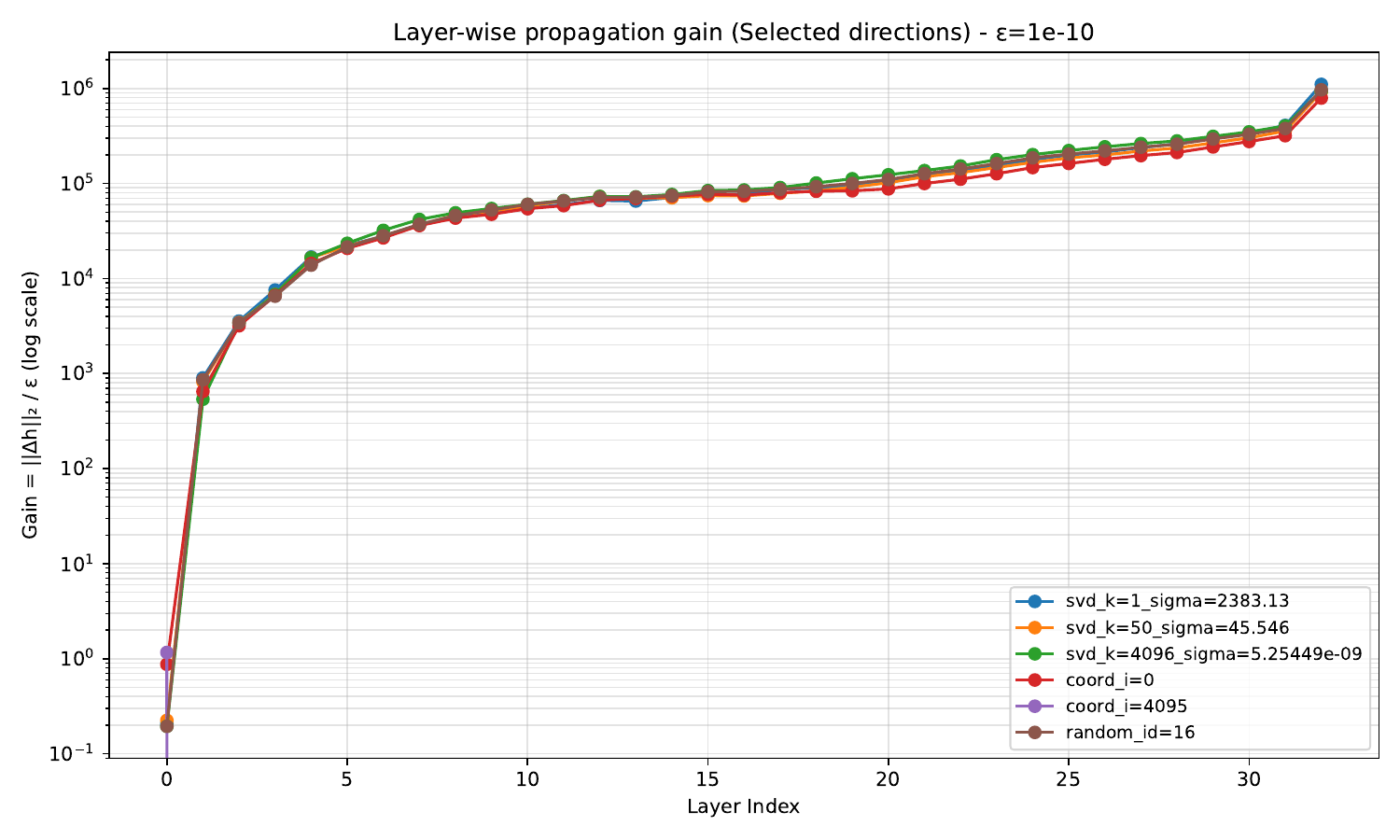}
    \caption{Layer-wise propagation profile at $\epsilon=10^{-10}$. Directional structure collapses: multiple singular directions, coordinate directions, and a random direction exhibit similar growth. This supports an instability-dominated regime where effective sensitivity is governed by scale rather than singular value.}
    \label{fig:layer_gain_small}
\end{figure}

Figure~\ref{fig:layer_gain_small} shows the same layer-wise plot for a much smaller perturbation ($\epsilon=10^{-10}$). Here the direction dependence largely collapses: singular directions with widely different $\sigma_k$, coordinate directions, and a random direction follow similar amplification trajectories and reach very large gains in later layers. This indicates an instability-dominated regime in which finite-precision effects and depth-wise amplification can dominate the directionality implied by the Jacobian spectrum.

Taken together, the overview and layer-wise results support an avalanche-like mechanism: once a perturbation survives rounding at the embedding interface, it can be amplified through depth in a way that becomes weakly dependent on the initial direction.

\subsection{Microscopic instability and constant-like plateaus across prompts}
We quantify how often microscopic perturbations cause noticeable representation changes and how this varies across datasets and models, revealing the prevalence of constant-like regions versus chaotic jumps.

We quantify local instability along a one-dimensional sweep in the embedding space aligned with a locally sensitive direction. For an ordered sweep $\epsilon_1 < \cdots < \epsilon_T$ and LPTs $m_i = M(x_{\epsilon_i,v})$, we define a finite-difference instability statistic
\begin{equation}
I_i = \frac{\|m_i - m_{i-1}\|_2}{\epsilon_i - \epsilon_{i-1}}, \quad i=2,\dots,T. \tag{5}
\end{equation}
We report the mean and median of $\{I_i\}$, along with the maximum drift $\max_i \|m_i - m_1\|_2$ over the sweep. We also report the logit margin at the last prompt position, defined as the difference between the largest and second-largest logits; we summarize it by its mean and minimum over the sweep.

Table~\ref{tab:instability_summary} shows a consistent spiky pattern under Float32 with a microscopic perturbation range: the median instability is 0.0 while the mean instability is orders of magnitude larger. This indicates that most consecutive perturbation steps produce no measurable change in the LPT (constant-like plateaus), but rare steps trigger discrete representation jumps whose finite-difference slopes become extremely large because the step size is tiny.

\begin{table}[t]
\centering
\caption{AGGREGATE STABILITY METRICS. \texttt{Llama-3.1-8B} RESULTS USE 100 PROMPTS PER DATASET (GPU). \texttt{GPT-OSS-20B} RESULTS USE 10 PROMPTS PER DATASET (CPU). MEDIAN INSTABILITY OF 0.0 INDICATES PREVALENT CONSTANT-LIKE PLATEAUS, WHILE LARGE MEAN INSTABILITY INDICATES RARE BUT LARGE FINITE-DIFFERENCE SPIKES.}
\label{tab:instability_summary}
\resizebox{\columnwidth}{!}{%
\begin{tabular}{llccccc}
\toprule
\textbf{Model} & \textbf{Dataset} & \textbf{Mean Inst.} & \textbf{Median Inst.} & \textbf{Max Drift} & \textbf{Mean Margin} & \textbf{Min Margin} \\
\midrule
GPT-OSS & AdvBench    & 6.76$\times 10^{5}$ & 0.0 & 2.92$\times 10^{-4}$ & 0.6196 & 0.6196 \\
GPT-OSS & TruthfulQA  & 2.63$\times 10^{5}$ & 0.0 & 3.43$\times 10^{-4}$ & 0.9767 & 0.9767 \\
\midrule
Llama-3.1 & AdvBench   & 3.78$\times 10^{6}$ & 0.0 & 1.02$\times 10^{-4}$ & 0.6175 & 0.6175 \\
Llama-3.1 & TruthfulQA & 8.84$\times 10^{6}$ & 0.0 & 1.46$\times 10^{-2}$ & 0.5351 & 0.5035 \\
\bottomrule
\end{tabular}%
}
\end{table}

\begin{figure}[t]
    \centering
    \includegraphics[width=0.92\columnwidth]{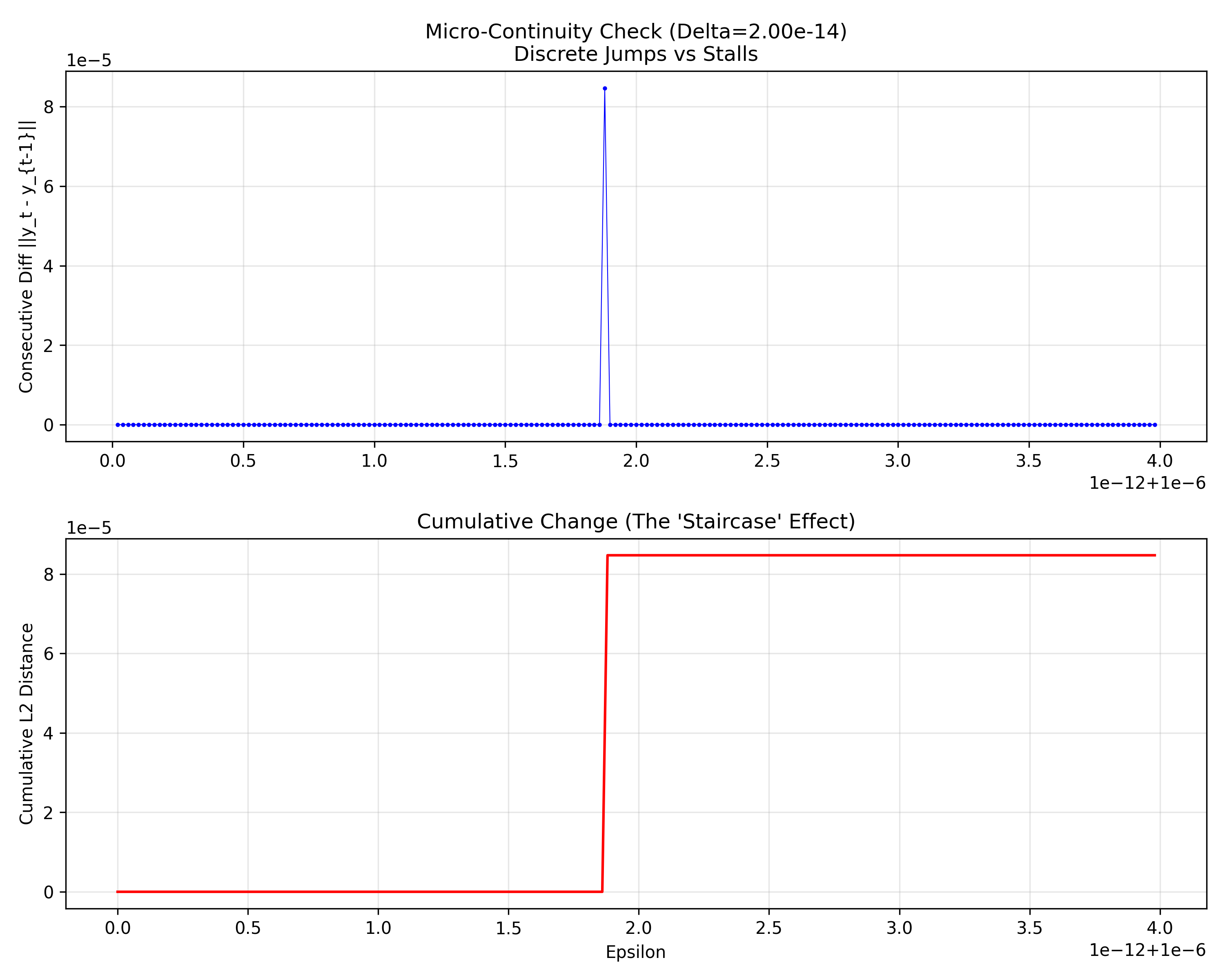}
    \caption{Micro-continuity analysis: consecutive differences exhibit long constant-like stalls with rare discrete jumps, producing a staircase-like cumulative change in the output representation.}
    \label{fig:micro_steps}
\end{figure}

\textbf{Micro-continuity at sub-ULP steps:}
To make the plateau-and-jump mechanism explicit, we probe $M(x)$ using consecutive perturbations separated by a $\Delta$ that is far smaller than typical Float32 spacing at the embedding magnitude. Figure~\ref{fig:micro_steps} shows long runs of consecutive differences indistinguishable from zero (stalls) punctuated by discrete jumps, producing a staircase cumulative change. This explains why median instability becomes exactly zero under tiny steps while the mean can remain very large: dividing a rare jump by a tiny $\Delta$ yields a huge finite-difference slope even when the absolute change in representation is small.

\subsection{Chaotic decision boundaries and regions}
We examine whether microscopic representation changes affect downstream outputs. They matter most when the model is near a tie between the top candidate tokens (i.e., when the logit margin is small). We construct near-tie scenarios and probe two-dimensional perturbation planes to visualize decision boundary geometry.

Starting from an embedding $x_0$ selected so that the top two logits are nearly equal ($L_1(x_0)\approx L_2(x_0)$), we perturb in two-dimensional planes spanned by right singular vectors: $x = x_0 + \epsilon_1 v_i + \epsilon_2 v_j$ with $\epsilon_1,\epsilon_2 \in [-10^{-8},10^{-8}]$ and step size $10^{-10}$. We test three planes: $(v_1,v_2)$, $(v_1,v_{10})$, and $(v_1,v_{4096})$, and record which of the top two tokens wins at each grid point.

\begin{figure}[t]
    \centering
    \subfloat[$v_1$-$v_2$ plane]{
        \includegraphics[width=0.45\columnwidth]{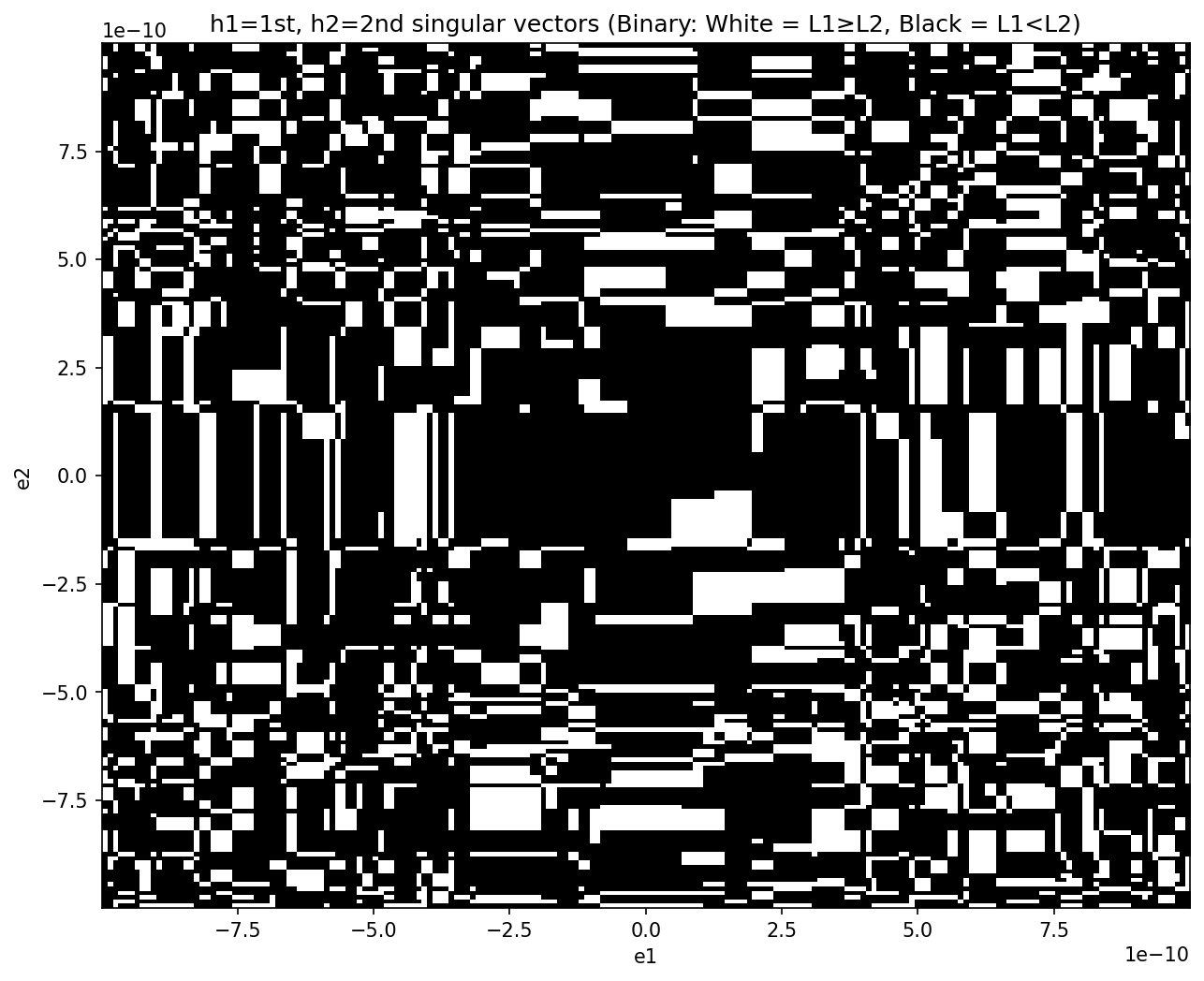}
        \label{fig:decision_1st_2nd}
    }\hfill
    \subfloat[$v_1$-$v_{4096}$ plane]{
        \includegraphics[width=0.45\columnwidth]{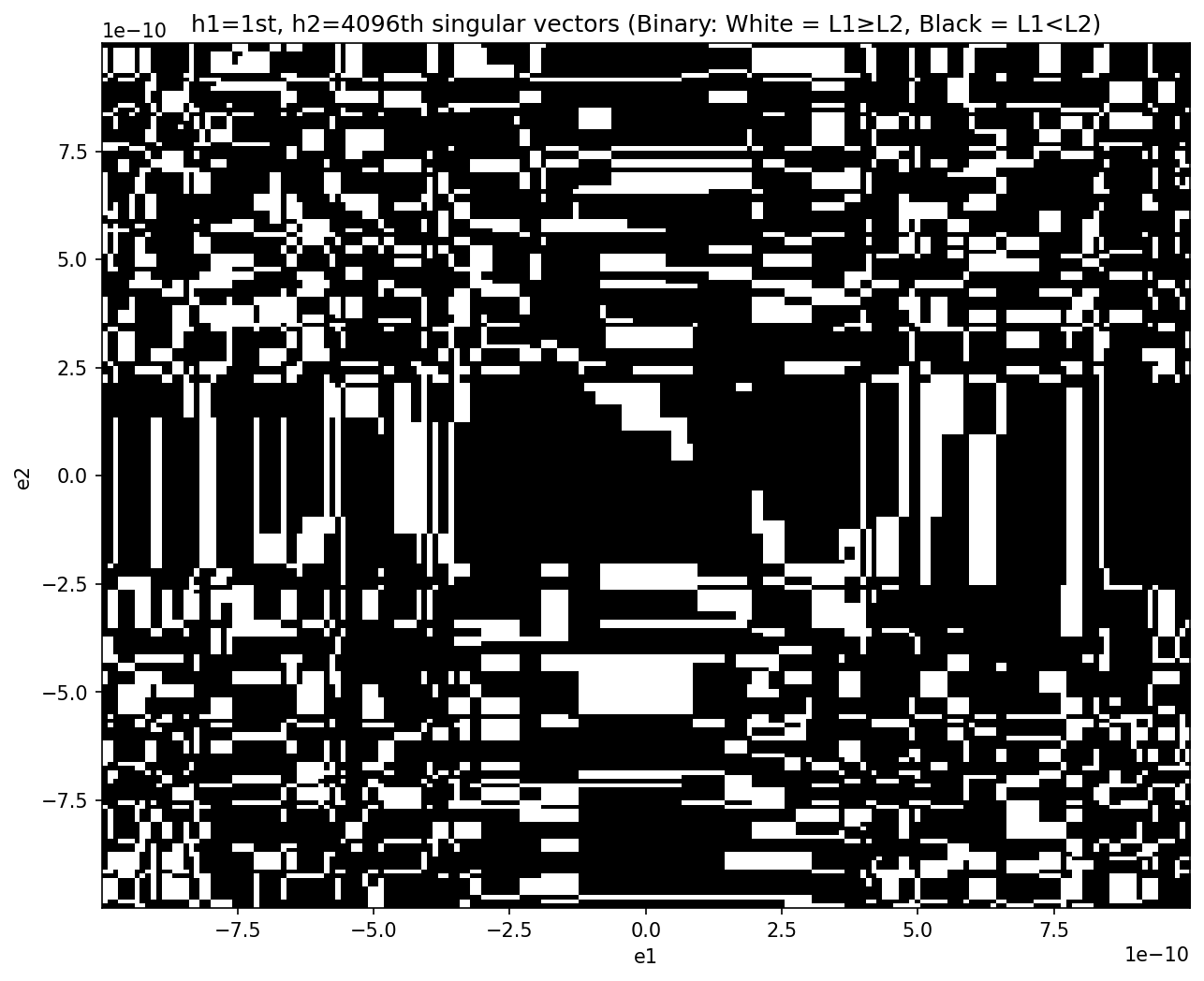}
        \label{fig:decision_1st_4096th}
    }
    \caption{Binary decision maps near a constructed near-tie point. White: $L_1 \geq L_2$ (Token 1 wins). Black: $L_1 < L_2$ (Token 2 wins). The fragmented regions indicate that near-tie decisions can flip under microscopic perturbations, including along low-$\sigma$ directions.}
    \label{fig:decision_boundaries}
\end{figure}

We quantify boundary irregularity using: (1) \textbf{Flip Frequency}, the fraction of adjacent grid cells with different predictions; (2) \textbf{Region Fragmentation}, the number of disconnected prediction regions; and (3) \textbf{Zero-Crossing Density}, the number of boundary crossings per line scan normalized by grid dimensions. Table~\ref{tab:decision_boundary_chaos} shows that all planes produce highly fragmented decision regions with dense boundary crossings. The same qualitative behavior persists even in the $(v_1,v_{4096})$ plane, indicating that near-tie decision sensitivity is not confined to a small subspace of ``high-$\sigma$'' directions. Figure~\ref{fig:decision_boundaries} visualizes this fragmentation through binary decision maps showing which token wins at each point in the perturbation grid. The salt-and-pepper patterns with no smooth regions reveal that microscopic perturbations cause erratic decision flips across the entire plane, with vertical patterns indicating rounding avalanches where cascading bit-flips trigger sudden output jumps.  The visualizations use the prompt \textit{"The capital of France is"}, chosen because its singular value spectrum ($\sigma_1 = 615.3$, $\sigma_2 = 386.9$, $\sigma_{4096} \approx 3 \times 10^{-9}$) represents typical input embeddings. Under normal conditions, the top token should have a clear lead over the second; however, after perturbation, the top two tokens exhibit nearly balanced logits ($L_1 \approx L_2$), creating ideal conditions for observing decision boundary chaos.

\begin{table}[t]
\centering
\caption{DECISION BOUNDARY IRREGULARITY METRICS ACROSS SINGULAR VECTOR PLANES NEAR A NEAR-TIE POINT $x_0$.}
\label{tab:decision_boundary_chaos}
\begin{tabular}{lccc}
\toprule
\textbf{Metric} & \textbf{$v_1$-$v_2$} & \textbf{$v_1$-$v_{10}$} & \textbf{$v_1$-$v_{4096}$} \\
\midrule
Flip Frequency (\%) & 16.14 & 16.29 & 15.42 \\
Fragmentation & 861 & 866 & 779 \\
Crossing Density & 51.0 & 52.5 & 49.0 \\
\bottomrule
\end{tabular}
\end{table}

\textbf{Directional Stability Mapping:} To systematically characterize instability across the embedding manifold, we measured maximum stable perturbation magnitudes using two complementary experiments: \textbf{(1)} angular sweeps in the $(v_1, v_2)$ plane for representation-level bit-flips, and \textbf{(2)} perturbations along all 4096 singular vectors.

\paragraph{Angular Stability in the $(v_1, v_2)$ Plane.}
We parameterized perturbations as $\delta = s(\cos\theta \cdot v_1 + \sin\theta \cdot v_2)$ with $\theta \in [0, 2\pi)$ sampled at 1000 angles. For each direction, we performed exponential search followed by binary search with ULP refinement to identify $s_{\max}(\theta)$, the maximum perturbation magnitude before representation bit-flip. This ULP-precise approach ensures we find the exact floating-point boundary where $M(x_0) = M(x_0 + s_{\max}\delta)$ but $M(x_0 + s_{\text{next}}\delta) \neq M(x_0)$, where $s_{\text{next}} = \texttt{nextafter}(s_{\max})$ in the float32 lattice.

\begin{figure}[htbp]
    \centering
    \begin{subfigure}[b]{0.48\textwidth}
        \includegraphics[width=\linewidth]{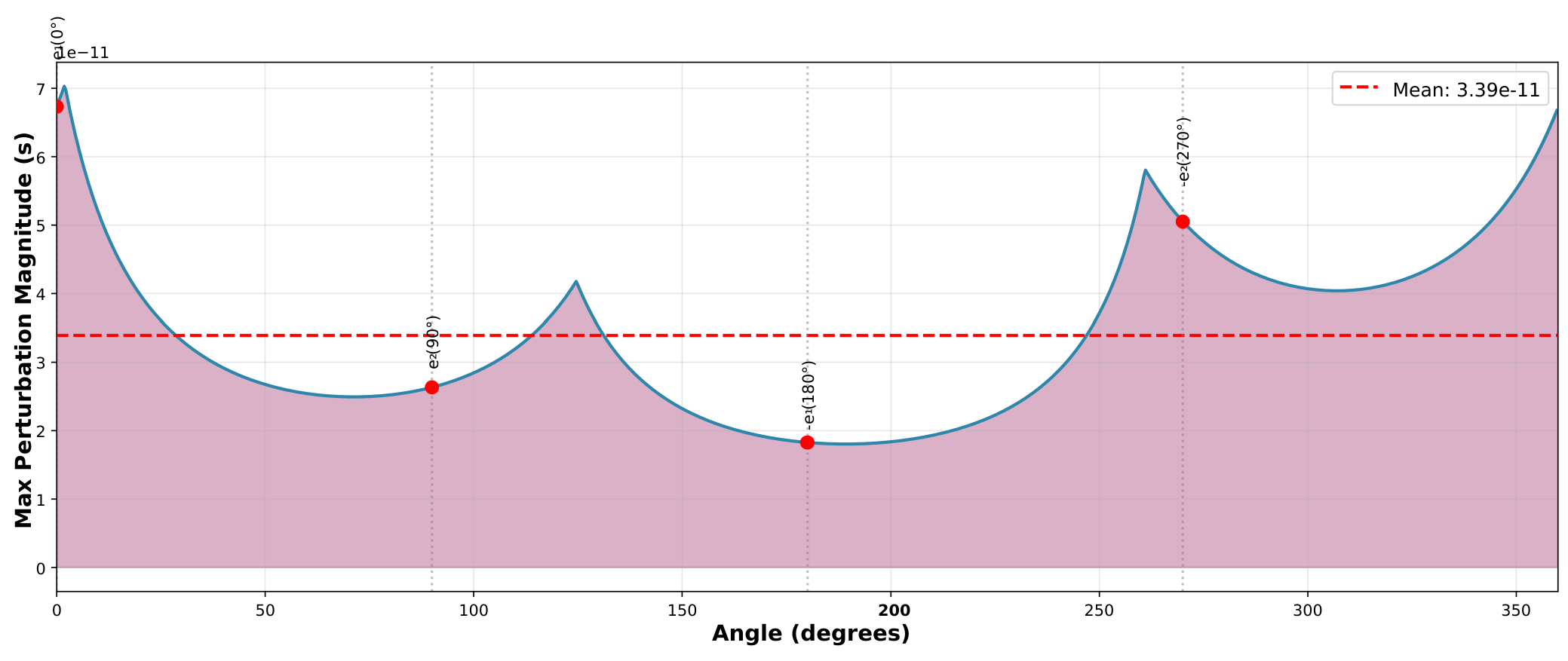}
        \caption{Angular stability profile}
        \label{fig:degrees_plot}
    \end{subfigure}
    \hfill
    \begin{subfigure}[b]{0.48\textwidth}
        \includegraphics[width=\linewidth]{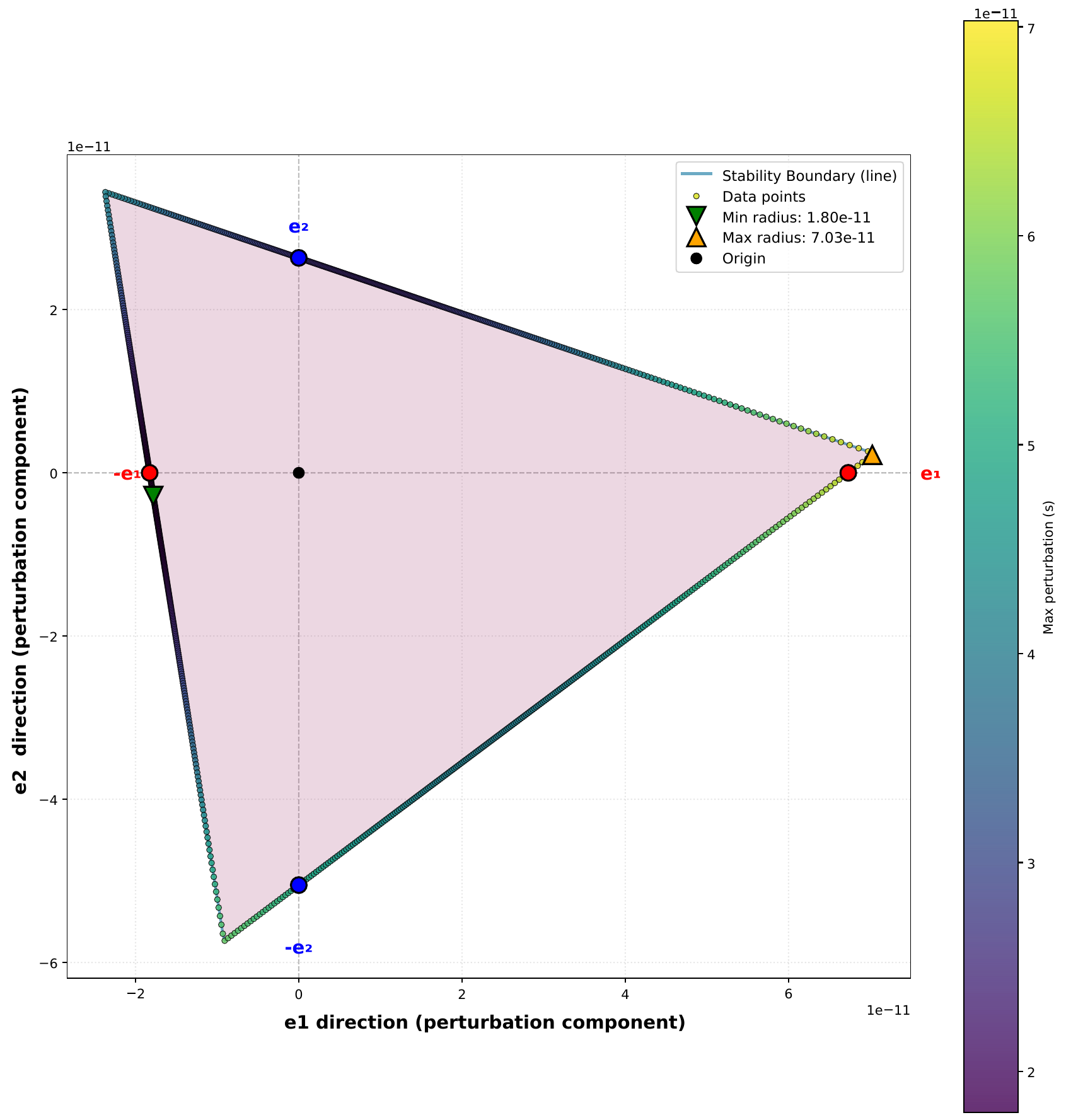}
        \caption{Boundary in $(v_1, v_2)$ space}
        \label{fig:cartesian_e1_e2}
    \end{subfigure}
    \caption{Directional stability in the $(v_1, v_2)$ plane. (a) Maximum stable perturbation $s_{\max}(\theta)$ varies by $4\times$ across angles. (b) Cartesian projection reveals asymmetric polygonal boundary. Geometric distortion demonstrates that stability is determined by input-dependent rounding dynamics, not by singular values.}
    \label{fig:angular_stability}
\end{figure}

Figure~\ref{fig:angular_stability} reveals extreme directional variability spanning four orders of magnitude. The angular profile shows $s_{\max}$ ranges from $1.80 \times 10^{-11}$ at $\theta \approx 180^{\circ}$ to $7.03 \times 10^{-11}$ at $\theta \approx 270^{\circ}$, with mean $3.39 \times 10^{-11}$. The Cartesian projection reveals an asymmetric polygonal boundary, not an ellipse as singular value theory would predict, but a distorted polygon with sharp vertices indicating catastrophic fragility in specific directions and extended lobes indicating Deep Constant Regions in others.

\begin{figure}[htbp]
    \centering
    \includegraphics[width=0.85\columnwidth]{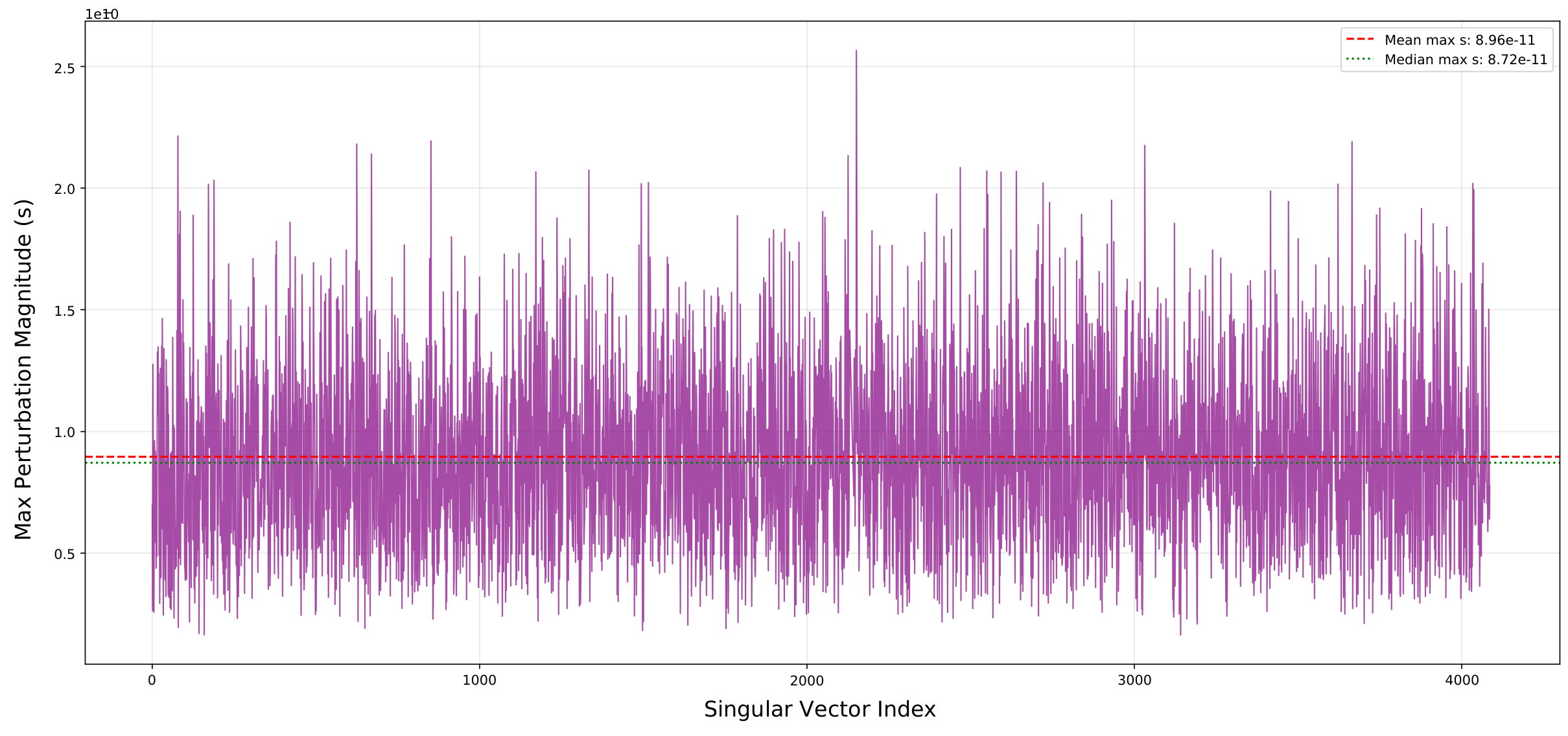}
    \caption{Maximum stable perturbation magnitude along all 4096 singular vectors. Despite singular values spanning $\sigma_1 = 615.3$ to $\sigma_{4096} \approx 0$, the stability boundary remains nearly constant ($s_{\max} \sim 10^{-10}$) across the entire spectrum. This universality proves instability pervades the entire embedding manifold.}
    \label{fig:singular_vector_stability}
\end{figure}

\paragraph{Universal Instability Across All Singular Vectors.}
To test whether instability is confined to high-sensitivity subspaces, we measured $s_{\max}$ along all 4096 singular vectors $v_i$ using identical binary search methodology. Figure~\ref{fig:singular_vector_stability} demonstrates that \textbf{instability is universal}: perturbations along $v_1$ ($\sigma_1 = 615.3$), $v_{50}$ ($\sigma_{50} = 31.1$), and $v_{4096}$ ($\sigma_{4096} \approx 0$) all exhibit $s_{\max} \sim 10^{-10}$, varying only by $3\times$ despite the singular values spanning from near-zero to over 600. The mean $s_{\max} = 8.96 \times 10^{-11}$ with standard deviation $4.12 \times 10^{-11}$ confirms that maximum stable perturbation is determined by floating-point quantization effects rather than spectral properties. Random spikes to $s_{\max} \sim 2.5 \times 10^{-10}$ occur sporadically across the entire singular spectrum, indicating that certain directions accidentally align with rounding plateaus, these correspond to the extended lobes observed in Figure~\ref{fig:cartesian_e1_e2}.

These experiments establish a critical finding: the chaotic decision boundaries are not artifacts of analyzing specific high-sensitivity planes, but reflect fundamental instability that \textbf{pervades the entire 4096-dimensional embedding space}. The fact that $s_{\max}$ varies by only $3\times$ across six orders of magnitude in singular values demonstrates that stability boundaries are determined by input-dependent rounding dynamics, not by the Jacobian's spectral structure.

\subsection{Effects of floating-point precision}
Assessing whether the three-regime picture is specific to Float32 or persists under other numeric formats shows how representational granularity affects the onset and location of instability-dominated behavior. We repeat directional sensitivity analysis under BFloat16 and Float64 precision.

\begin{figure}[t]
    \centering
    \subfloat[BFloat16]{
        \includegraphics[width=0.92\columnwidth,
        trim=0 0 0 25,
        clip
        ]{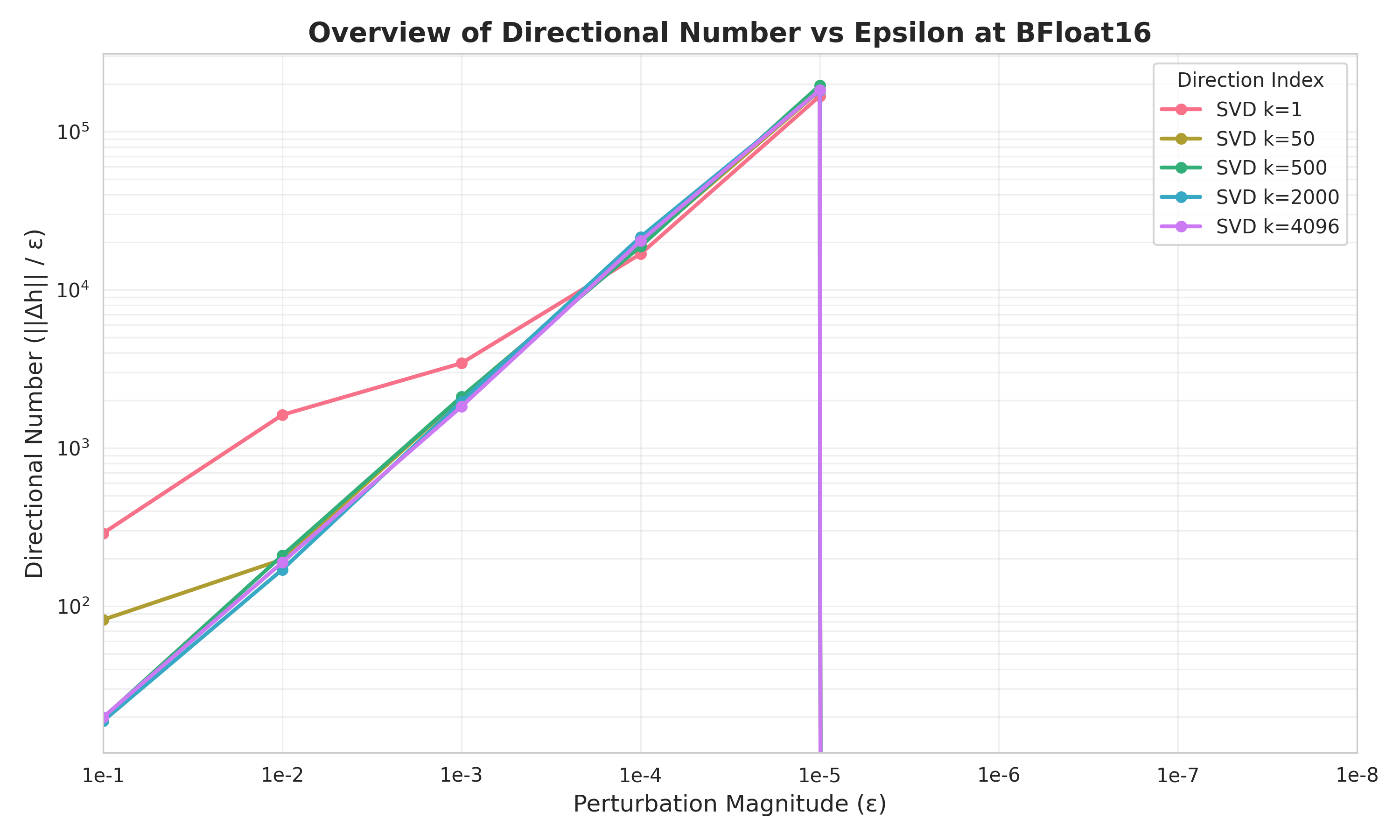}
        \label{fig:directional_overview_bf16}
    }\hfill
    \subfloat[Float64]{
        \includegraphics[width=0.92\columnwidth,
        trim=0 0 0 25,
        clip
        ]{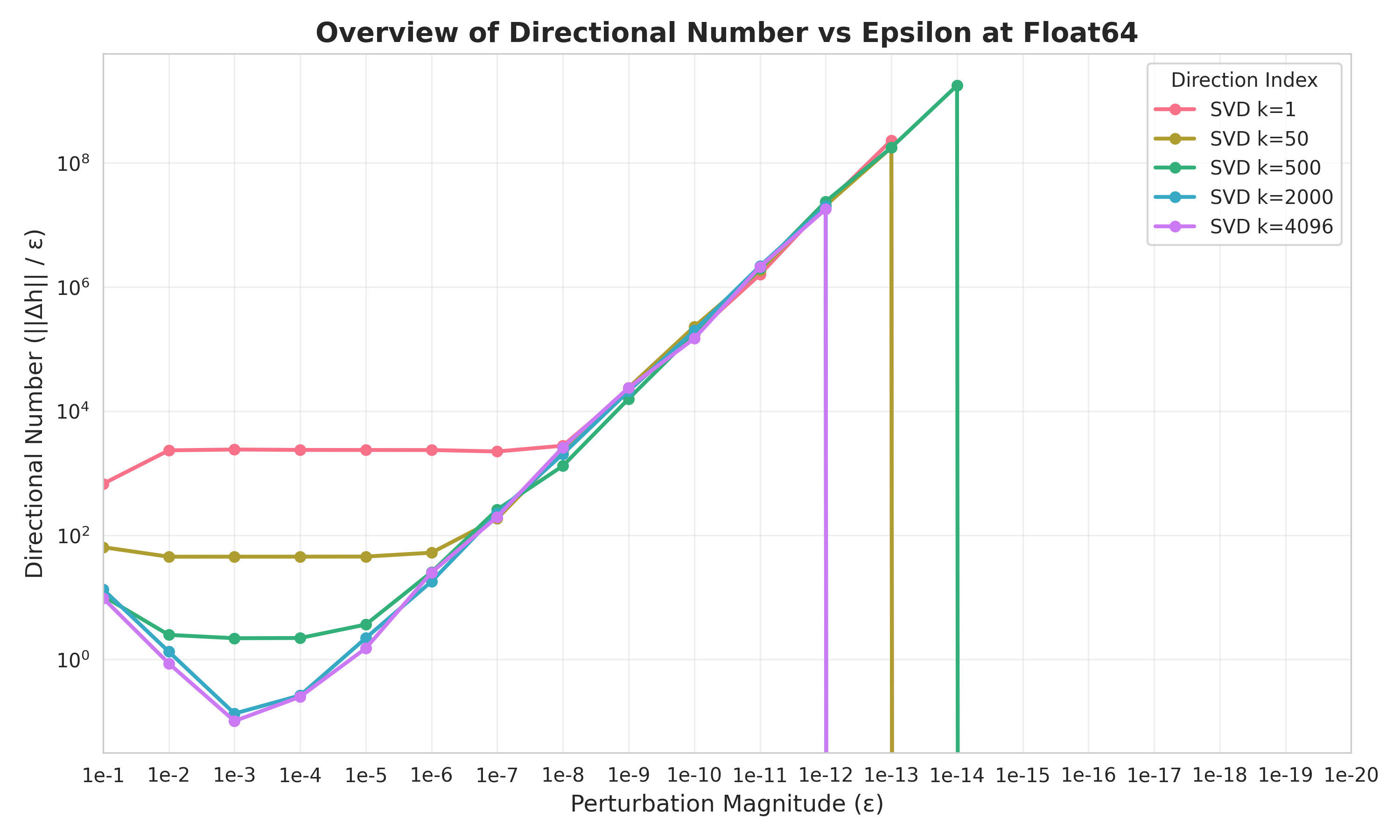}
        \label{fig:directional_overview_fp64}
    }
    \caption{Directional number vs.\ $\epsilon$ at layer index 32 under different floating-point precisions. Precision shifts the $\epsilon$ ranges where plateau behavior and rapid growth appear, while preserving the overall scale dependence.}
    \label{fig:directional_overview_bf16_fp64}
\end{figure}

Figure~\ref{fig:directional_overview_bf16_fp64} repeats the directional-number sweep at the same layer index using BFloat16 and Float64. The qualitative structure remains: there is a regime where the curves partially collapse across directions and a regime where directionality becomes more visible at larger perturbations. What changes with precision is the \emph{location} of the transition scales. In BFloat16, the onset of plateau behavior and the transition to large effective directional numbers occurs at coarser $\epsilon$ because representational spacing is larger. In Float64, the same effects are shifted toward much smaller $\epsilon$, consistent with a finer representational grid and a wider range of distinguishable perturbations before stalling dominates.

These results support the claim that the regimes are not tied to a single floating-point type. Precision changes how quickly perturbations are rounded away and where the instability-dominated behavior becomes visible, but it does not remove the scale dependence of effective sensitivity when probing finite-$\epsilon$ behavior \cite{goldberg1991floating} \cite{8766229}.

\subsection{Mitigation via Noise Averaging}

Since the root cause of instability is the propagation of floating-point 
rounding errors through deep networks, we propose a simple but effective mitigation strategy: averaging multiple forward passes with injected noise. For a perturbation $x_0 + \epsilon \cdot s_i$ along singular vector $s_i$, 
instead of computing a single absolute directional condition number estimate $\kappa_{abs} = \|f(x_0 + \epsilon s_i) - f(x_0)\| / \epsilon$, we compute a smoothed estimate by averaging $n$ noisy evaluations: $\kappa_{\text{smooth}} = \|\frac{1}{n}\sum_{j=1}^{n} f(x_0 + \epsilon s_i + v_j) - 
\frac{1}{n}\sum_{j=1}^{n} f(x_0 + v_j)\| / \epsilon$, where $v_j$ are small random perturbations along $s_i$ with magnitude $10^{-9}$. The key insight is that rounding errors are non-deterministic and vary across different computational paths, while the true model sensitivity remains constant. By the law of large numbers, averaging over multiple samples cancels out the stochastic rounding noise, revealing the underlying algorithmic sensitivity.

\begin{figure}[htbp]
    \centering
    \includegraphics[width=0.85\columnwidth]{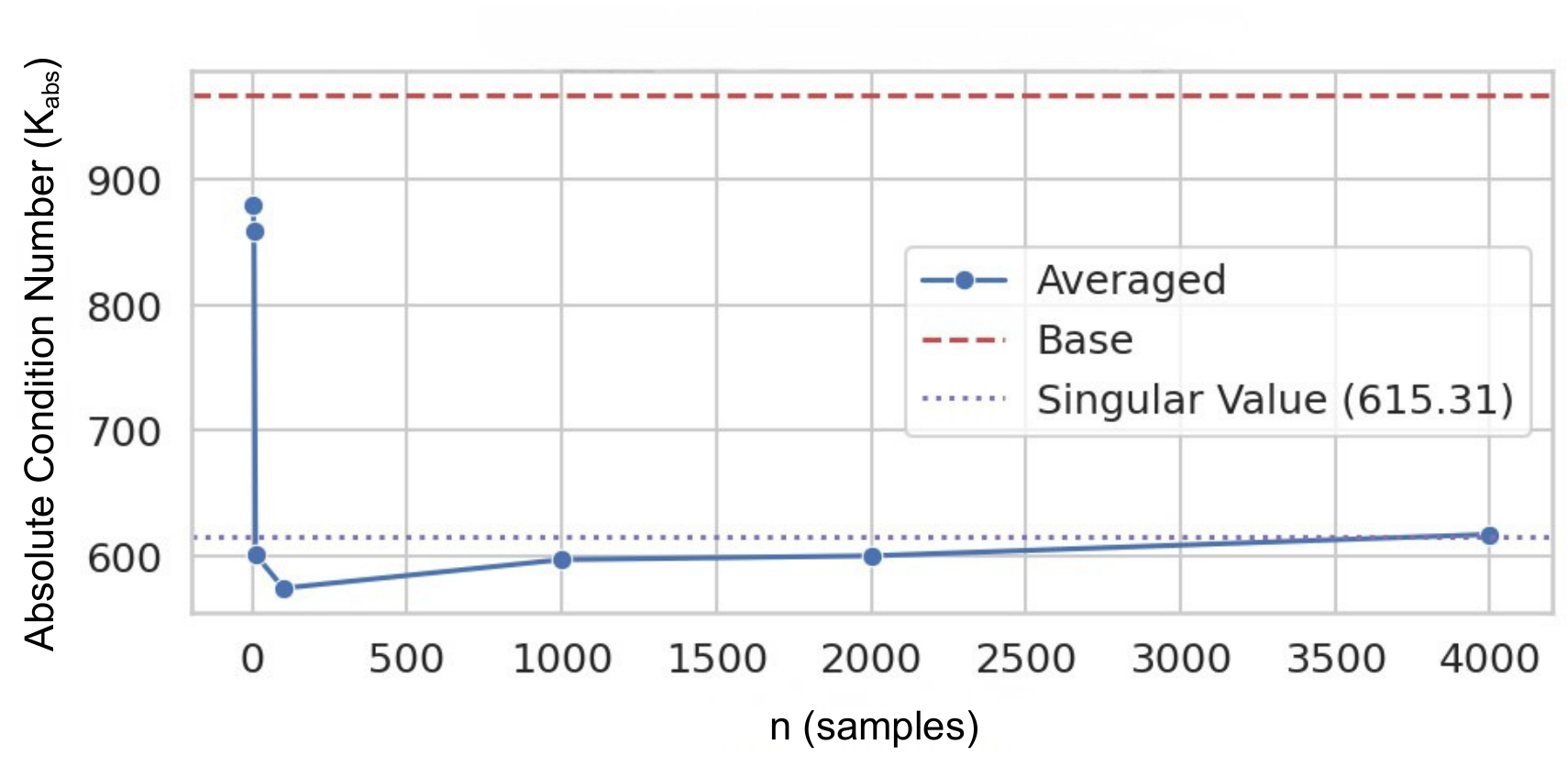}
    \caption{Absolute directional condition number convergence via noise averaging. 
The dashed red line shows the unaveraged baseline, while the blue curve shows 
the averaged estimate stabilizing at $\sim$600 by $n=100$ samples, converging 
to the theoretical singular value (dotted line, 615.31). Averaging eliminates 
rounding-error artifacts and recovers true model sensitivity.}
    \label{fig:mitigation_averaging}
\end{figure}

Figure~\ref{fig:mitigation_averaging} demonstrates the effectiveness of this approach. Without averaging ($n=1$), the empirical absolute directional condition number 
is above 900, exceeding the theoretical maximum singular value of 615.31 due to rounding error amplification. With just $n=10$ samples, the estimate drops sharply to $\sim$600, and by $n=100$, it stabilizes close to 600, converging to the true singular value. Further increases to $n=4000$ provide minimal additional benefit, confirming convergence. This mitigation transforms the chaotic, unreliable single-sample estimate into a stable, reproducible measurement that accurately reflects genuine model sensitivity rather than numerical artifacts. The method's practicality is demonstrated by its rapid convergence, even modest averaging ($n=10$-$100$) suffices to recover reliable estimates at negligible computational cost.

\section{Discussion}
Our findings show that LLMs operate at the boundary of numerical chaos, where perturbations exhibit binary outcomes: complete attenuation in Constant Regions or explosive amplification in Chaotic Regions. Directional sensitivity is scale-driven rather than spectrum-driven, and perturbations across all singular directions (spanning five orders of magnitude in singular values) exhibit nearly identical stability thresholds, contradicting classical conditioning theory. Instability originates from early-layer avalanche effects where microscopic errors cascade through depth, producing median instability of zero but extreme mean values, confirming discrete jumps rather than smooth divergence.

Near decision boundaries, microscopic perturbations fragment output space into hundreds of disconnected regions with crossing densities exceeding smooth expectations by 50×. This fractal geometry pervades the entire 4096-dimensional embedding space. For multi-agent systems, non-deterministic floating-point operations cause identical inputs to traverse different computational paths, explaining 23-31\% failure rates. Architectural differences affect stability: GPT-OSS-20B maintains larger Constant Regions and higher logit margins than Llama-3.1, suggesting normalization schemes significantly impact vulnerability. Increasing precision merely shifts regime boundaries without eliminating chaos, distributed systems cannot guarantee reproducibility due to non-associative reductions.

\textbf{Limitations and Future Work.} Our analysis focuses on inference-time instability. Future work should investigate: (1) whether training implicitly navigates toward stable regions; (2) architectural modifications to expand Constant Regions; (3) runtime boundary detection for adaptive precision; (4) connections between numerical chaos and adversarial vulnerability; and (5) extension to multimodal models.

\section{Conclusion}
We demonstrate that LLMs exhibit universal scale-dependent chaotic behaviors arising from floating-point rounding errors amplified through transformer depth. We identify three operating regimes: Constant (bitwise-frozen outputs), Chaotic (rounding-error dominated), and Signal-Dominated (input variations override noise). Directional sensitivity is determined by perturbation scale $\epsilon$ rather than Jacobian spectrum, all singular directions exhibit $s_{\max} \sim 10^{-10}$ despite five orders of magnitude variation in singular values. Instability originates from early-layer avalanche effects, and near decision boundaries fragments output space into hundreds of chaotic regions. These findings establish numerical stability as a fundamental constraint on LLM reproducibility across heterogeneous deployments, providing practitioners with a stability framework for safety-critical applications.


\bibliographystyle{ieeetr}
\bibliography{references}
\clearpage
\end{document}